\documentclass[11pt]{article}
\usepackage{acl2017}
\usepackage{times}
\usepackage{latexsym}
\usepackage{xspace}
\usepackage{arydshln}
\usepackage{graphicx}
\usepackage{multirow}
\usepackage{amsmath}
\usepackage{relsize}
\usepackage{url}
\usepackage{caption}
\usepackage{subcaption}
\usepackage{adjustbox,booktabs}
\usepackage{amsfonts,amssymb}
\usepackage{algorithm,algpseudocode}
\usepackage{hyperref}

\newcommand{\method}[1]{\texttt{#1}\xspace}
\newcommand{\lstm}{\method{vanilla-lstm}}
\newcommand{\lclm}{\method{lclm}}
\newcommand{\lstmlda}{\method{lstm$+$lda}}
\newcommand{\tdlm}{\method{tdlm}}
\newcommand{\lda}{\method{lda}}
\newcommand{\ntm}{\method{ntm}}
\newcommand{\tdlms}{\method{tdlm-small}}
\newcommand{\tdlml}{\method{tdlm-large}}
\newcommand{\wordtovec}{\method{word2vec}}

\newcommand{\dataset}[1]{\textsc{#1}\xspace}
\newcommand{\apnews}{\dataset{apnews}}
\newcommand{\imdb}{\dataset{imdb}}
\newcommand{\bnc}{\dataset{bnc}}
\newcommand{\tnews}{\dataset{20news}}

\newcommand{\smallurl}[1]{{\smaller{\url{#1}}}}

\newcommand{\secref}[2][]{Section#1~\ref{sec:#2}}

\newcommand{\tabref}[2][]{Table#1~\ref{tab:#2}}
\newcommand{\figref}[2][]{Figure#1~\ref{fig:#2}}

\newcommand{\eqnref}[2][]{Equation#1~(\ref{eqn:#2})}

\newcommand{\unk}{$\langle$unk$\rangle$\xspace}
\newcommand{\z}{\phantom{0}}

\newcommand\email{\begingroup \urlstyle{tt}\smaller\Url}

\aclfinalcopy 
\def\aclpaperid{265}

\title{Topically Driven Neural Language Model}

\author{Jey Han Lau$^{1,2}$ \qquad Timothy Baldwin$^{2}$ \qquad Trevor 
Cohn$^{2}$ \\[1ex]
    $^1$ IBM Research \\[1ex]
    $^2$ School of Computing and Information Systems,\\The University of
Melbourne \\
    \email{jeyhan.lau@gmail.com}, \email{tb@ldwin.net}, 
\email{t.cohn@unimelb.edu.au}}

\date{}

\begin{document}

\maketitle

\begin{abstract}
  Language models are typically applied at the sentence level, without
  access to the broader document context.  We present a neural language
  model that incorporates document context in the form of a topic 
  model-like architecture, thus providing a succinct representation of 
the
  broader document context outside of the current sentence.  Experiments
  over a range of datasets demonstrate that our model outperforms a pure
  sentence-based model in terms of language model perplexity, and leads
  to topics that are potentially more coherent than those produced by a 
  standard LDA topic model.  Our model also has the ability to generate 
related sentences for a topic, providing another way to interpret 
topics.


\end{abstract}

\section{Introduction}

Topic models provide a powerful tool for extracting the macro-level
content structure of a document collection in the form of the latent
topics (usually in the form of multinomial distributions over terms),
with a plethora of applications in NLP
\citep{Hall+:2008,Newman+:2009a,Wang:McCallum:2006}. A myriad of
variants of the classical LDA method \citep{Blei+:2003} have been
proposed, including recent work on neural topic models
\citep{Cao+:2015,Wan+:2012,Larochelle:Lauly:2012,Hinton+:2009}.

Separately, language models have long been a foundational component of
any NLP task involving generation or textual normalisation of a noisy
input (including speech, OCR and the processing of social media
text). The primary purpose of a language model is to predict the
probability of a span of text, traditionally at the sentence level, under
the assumption that sentences are independent of one another, although
recent work has started using broader local context such as the
preceding sentences \cite{Wang+:2016,Ji+:2016}.

In this paper, we combine the benefits of a topic model and language
model in proposing a topically-driven language model, whereby we jointly
learn topics and word sequence information. This allows us to both
sensitise the predictions of the language model to the larger document
narrative using topics, and to generate topics which are better
sensitised to local context and are hence more coherent and
interpretable.

Our model has two components: a language model and a
topic model. We implement both components using neural networks, and
train them jointly by treating each component as a sub-task in a
multi-task learning setting.  We show that our model is superior to
other language models that leverage additional context, and that the
generated topics are potentially more coherent than LDA topics.  The 
architecture of the model provides an extra dimensionality of topic 
interpretability, in supporting the generation of sentences from a topic 
(or mix of
topics). It is also highly flexible, in its ability to be supervised and
incorporate side information, which we show to further improve language
model performance. An open source implementation of our model is
available at: 
\url{https://github.com/jhlau/topically-driven-language-model}.


\begin{figure*}[t]
\begin{center}
\includegraphics[width=1.0\textwidth]{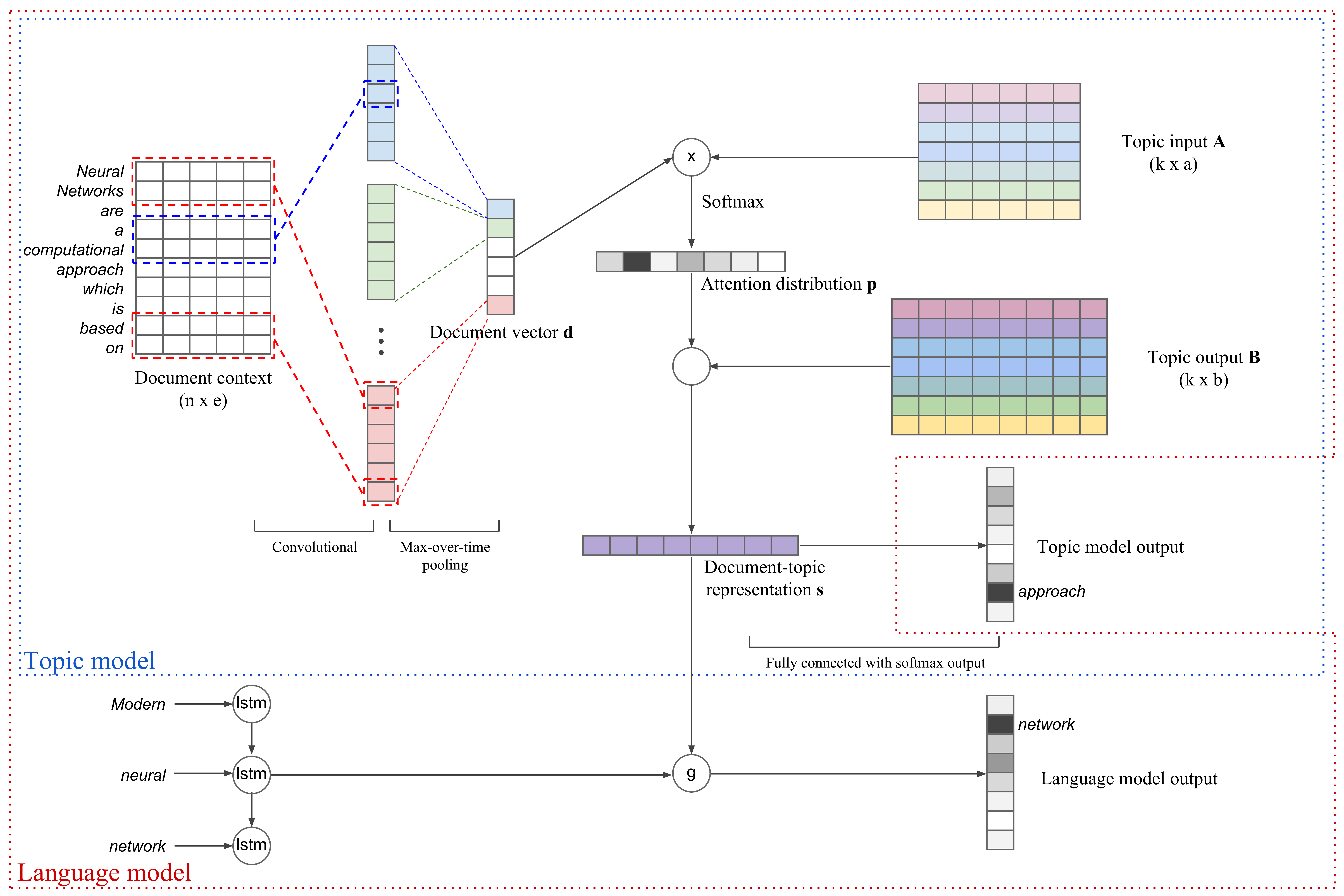}
\end{center}
\caption{Architecture of \tdlm. Scope of the models are denoted by 
dotted lines: blue line denotes the scope of the topic model, red the 
language model.}
\label{fig:model}
\end{figure*}

\section{Related Work}

\citet{Griffiths+:2004} propose a model that learns topics and word 
dependencies using a Bayesian framework. Word generation is driven by 
either LDA or an HMM. For LDA, a word is generated based on a sampled topic 
in the document.  For the HMM, a word is conditioned on previous words. A 
key difference over our model is that their language model is driven by an
HMM, which uses a fixed window and is therefore unable to track long-range dependencies.

\citet{Cao+:2015} relate the topic model view of documents and words
--- documents having a multinomial distribution over topics and topics
having a multinomial distributional over words  --- from a neural
network perspective by embedding these relationships in differentiable
functions. With that, the model lost the stochasticity and Bayesian
inference of LDA but gained non-linear complex representations. The
authors further propose extensions to the model to do supervised
learning where document labels are given.


\citet{Wang+:2016} and \citet{Ji+:2016} relax the sentence independence assumption in 
language modelling, and use preceeding sentences as additional context.  
By treating words in preceeding sentences as a bag of words, 
\citet{Wang+:2016} use an attentional mechanism to focus on these words 
when predicting the next word. The authors show that the incorporation 
of additional context helps language models.

\section{Architecture}

The architecture of the proposed topically-driven language model
(henceforth ``\tdlm'') is illustrated in \figref{model}.  There are two
components in \tdlm: a language model and a topic model. The language
model is designed to capture word relations in sentences, while the
topic model learns topical information in documents. The topic model
works like an auto-encoder, where it is given the document words as
input and optimised to predict them.

The topic model takes in word embeddings of a document and generates a 
document vector using a convolutional network.  Given the document 
vector, we associate it with the topics via an attention scheme
to compute a weighted mean of topic vectors, which is then used to 
predict a word in the document.

The language model is a standard LSTM language model 
\citep{Hochreiter+:1997,Mikolov+:2010}, but it incorporates the weighted 
topic vector generated by the topic model to predict succeeding words.

Marrying the language and topic models allows the language model to be 
topically driven, i.e.\ it models not just word contexts but also 
the document context where the sentence occurs, in the form of topics.

\subsection{Topic Model Component}
\label{sec:tmc}

Let $\mathbf{x}_i \in \mathbb{R}^e$ be the $e$-dimensional word vector 
for the $i$-th word in the document. A document of $n$ words is 
represented as a concatenation of its word vectors:
\begin{equation*}
\mathbf{x}_{1:n} = \mathbf{x}_1 \oplus \mathbf{x}_2 \oplus ...  \oplus 
\mathbf{x}_n
\end{equation*}
where $\oplus$ denotes the concatenation operator. We use a number of 
convolutional filters to process the word vectors, but for clarity we 
will explain the network with one filter.

Let $\mathbf{w}_v \in \mathbb{R}^{eh}$ be a convolutional filter which 
we apply to a window of $h$ words to generate a feature.  A feature 
$c_i$ for a window of words
$\mathbf{x}_{i:i+h-1}$ is given as follows:
\begin{equation*}
    c_i = I(\mathbf{w}^\intercal_v\mathbf{x}_{i:i+h-1} + b_v)
\end{equation*}
where $b_v$ is a bias term and $I$ is the identity function.\footnote{A
non-linear function is typically used here, but preliminary experiments
suggest that the identity function works best for \tdlm.} A feature map
$\mathbf{c}$ is a collection of features computed from all windows of
words:
\begin{equation*}
\mathbf{c} = [c_1, c_2, ..., c_{n-h+1}]
\end{equation*}
where $\mathbf{c} \in \mathbb{R}^{n-h+1}$. To capture the most salient 
features in $\mathbf{c}$, we apply a max-over-time pooling operation
\citep{Collobert+:2011}, yielding a scalar:
\begin{equation*}
d = \max_i c_i
\end{equation*}

In the case where we use $a$ filters, we have $\mathbf{d} \in 
\mathbb{R}^a$, and this constitutes the vector representation of the 
document generated by the convolutional and max-over-time pooling 
network.

The topic vectors are stored in two lookup tables $\mathbf{A} \in 
\mathbb{R}^{k \times a}$ (input vector) and $\mathbf{B} \in 
\mathbb{R}^{k \times b}$ (output vector), where $k$ is the number of 
topics, and $a$ and $b$ are the dimensions of the topic vectors.

To align the document vector $\mathbf{d}$ with the topics, we compute an
attention vector which is used to compute a document-topic
representation:\footnote{The attention mechanism was
  inspired by memory networks
  \citep{Graves+:2014,Weston+:2014,Sukhbaatar+:2015,Tran+:2016}.  We
  explored various attention styles (including traditional schemes which
  use one vector for a topic), but found this approach to work best.}
\begin{align}
 \mathbf{p} &= \text{softmax}(\mathbf{A}\mathbf{d}) \label{eqn:attn} \\
 \mathbf{s} &= \mathbf{B}^\intercal \mathbf{p}
\label{eqn:hiddenreps}
\end{align}
where $\mathbf{p} \in \mathbb{R}^k$ and $\mathbf{s} \in \mathbb{R}^b$.  
Intuitively, $\mathbf{s}$ is a weighted mean of topic vectors, with the 
weighting given by the attention $\mathbf{p}$. This is inspired by the 
generative process of LDA, whereby documents are defined as having a 
multinomial distribution over topics.

Finally $\mathbf{s}$ is connected to a dense layer with softmax output 
to predict each word in the document, where each word is generated independently 
as a unigram bag-of-words, and the model is optimised using categorical 
cross-entropy loss.  In practice, to improve efficiency we compute loss 
for predicting a sequence of $m_1$ words in the document, where $m_1$ is 
a hyper-parameter.

\subsection{Language Model Component}

The language model is implemented using LSTM units 
\citep{Hochreiter+:1997}:
\begin{equation*}
\begin{split}
    \mathbf{i}_t &= \sigma(\mathbf{W}_i \mathbf{v}_t + \mathbf{U}_i 
\mathbf{h}_{t-1} + \mathbf{b}_i) \\
    \mathbf{f}_t &= \sigma(\mathbf{W}_f \mathbf{v}_t + \mathbf{U}_f 
\mathbf{h}_{t-1} + \mathbf{b}_f) \\
    \mathbf{o}_t &= \sigma(\mathbf{W}_o \mathbf{v}_t + \mathbf{U}_o 
\mathbf{h}_{t-1} + \mathbf{b}_i) \\
    \hat{\mathbf{c}}_t &= \text{tanh}(\mathbf{W}_c \mathbf{v}_t + 
\mathbf{U}_c \mathbf{h}_{t-1} + \mathbf{b}_c) \\
    \mathbf{c}_t &= \mathbf{f}_t \odot \mathbf{c}_{t-1} + \mathbf{i}_t 
\odot \hat{\mathbf{c}}_t \\
    \mathbf{h}_t &= \mathbf{o}_t \odot \text{tanh}(\mathbf{c}_t)
\end{split}
\end{equation*}
where $\odot$ denotes element-wise product; $\mathbf{i}_t$, 
$\mathbf{f}_t$, $\mathbf{o}_t$ are the input, forget and output 
activations respectively at time step $t$; and $\mathbf{v}_t$, 
$\mathbf{h}_t$ and $\mathbf{c}_t$ are the input word embedding, LSTM 
hidden state, and cell state, respectively. Hereinafter $\mathbf{W}$, 
$\mathbf{U}$ and $\mathbf{b}$ are used to refer to the model parameters.

Traditionally, a language model operates at the sentence level, 
predicting the next word given its history of words in the sentence.  
The language model of \tdlm incorporates topical information by 
assimilating the document-topic representation ($\mathbf{s}$) with the 
hidden output of the LSTM ($\mathbf{h}_t$) at each time step $t$.  To 
prevent \tdlm from memorising the next word via the topic model network, 
we exclude the current sentence from the document context.

We use a gating unit similar to a GRU \citep{Cho+:2014,Chung+:2014} to
allow \tdlm to learn the degree of influence of topical information on the language model:
\begin{equation}
\begin{split}
    \mathbf{z}_t &= \sigma(\mathbf{W}_z \mathbf{s} + \mathbf{U}_z 
\mathbf{h}_t + \mathbf{b}_z) \\
    \mathbf{r}_t &= \sigma(\mathbf{W}_r \mathbf{s} + \mathbf{U}_r 
\mathbf{h}_t + \mathbf{b}_r) \\
    \hat{\mathbf{h}}_t &= \text{tanh}(\mathbf{W}_h \mathbf{s} + 
\mathbf{U}_h (\mathbf{r}_t \odot \mathbf{h}_t) + \mathbf{b}_h) \\
    \mathbf{h}'_t &= (1 - \mathbf{z}_t) \odot \mathbf{h}_t + 
\mathbf{z}_t \odot \hat{\mathbf{h}}_t
\end{split}
\label{eqn:incorporatetm}
\end{equation}
where $\mathbf{z}_t$ and $\mathbf{r}_t$ are the update and reset gate 
activations respectively at timestep $t$. The new hidden state 
$\mathbf{h}'_t$ is connected to a dense layer with linear transformation 
and softmax output to predict the next word, and the model is optimised 
using standard categorical cross-entropy loss.

\subsection{Training and Regularisation}

\tdlm is trained using minibatches and SGD.\footnote{We use Adam as the 
optimiser \citep{Kingma+:2014}.} For the language model, a minibatch 
consists of a batch of sentences, while for the topic model it is a 
batch of documents (each predicting a sequence of $m_1$ words).

We treat the language and topic models as sub-tasks in a multi-task 
learning setting, and train them jointly using categorical cross-entropy 
loss. Most parameters in the topic model are shared by the language 
model, as illustrated by their scopes (dotted lines) in \figref{model}.


Hyper-parameters of \tdlm are detailed in \tabref{hyperparameters}. Word 
embeddings for the topic model and language model components are not 
shared, although their dimensions are the same ($e$).\footnote{Word 
embeddings are updated during training.}  For $m_1$, $m_2$ and $m_3$, 
sequences/documents shorter than these thresholds are padded.  Sentences 
longer than $m_2$ are broken into multiple sequences, and documents 
longer than $m_3$ are truncated.  Optimal hyper-parameter settings are 
tuned using the development set; the presented values are used for 
experiments in \secref[s]{lm} and \ref{sec:tm}.

\begin{table*}[t]
\begin{center}
\begin{adjustbox}{max width=0.85\textwidth}
\begin{tabular}{ccp{11cm}}
\toprule
\textbf{Hyper-} & \multirow{2}{*}{\textbf{Value}} & 
\multirow{2}{*}{\textbf{Description}} \\
\textbf{parameter} & \\
\midrule
$m_1$ & 3 & Output sequence length for topic model\\
{$m_2$} & 30 & Sequence length for language model \\
$m_3$ & 300,150,500 &  Maximum document length\\
$n_\mathit{batch}$ & 64 & Minibatch size \\
$n_\mathit{layer}$ & 1,2 & Number of LSTM layers\\
$n_\mathit{hidden}$ & 600,900 & LSTM hidden size \\
$n_\mathit{epoch}$ & 10 & Number of training epochs \\
$k$ & 100,150,200 & Number of topics \\
$e$ & 300 & Word embedding size \\
$h$ & 2 & Convolutional filter width \\
$a$ & 20 & Topic input vector size or number of features for 
convolutional filter\\
$b$ & 50 & Topic output vector size\\
$l$ & 0.001 & Learning rate of optimiser \\
$p_1$ & 0.4 & Topic model dropout keep probability \\
$p_2$ & 0.6 & Language model dropout keep probability \\
\bottomrule
\end{tabular}
\end{adjustbox}
\end{center}
\caption{\tdlm hyper-parameters; we experiment with 2 LSTM settings and 
3 topic numbers, and $m_3$ varies across the three domains (\apnews, \imdb, and 
  \bnc).}
\label{tab:hyperparameters}
\end{table*}

To regularise \tdlm, we use dropout regularisation 
\citep{Srivastava+:2014}.  We apply dropout to $\mathbf{d}$ and 
$\mathbf{s}$ in the topic model, and to the input word embedding and 
hidden output of the LSTM in the language model 
\citep{Pham+:2013,Zaremba+:2014}.

\section{Language Model Evaluation}
\label{sec:lm}

We use standard language model perplexity as the evaluation metric. In
terms of dataset, we use document collections from 3 sources: \apnews,
\imdb and \bnc. \apnews is a collection of Associated
Press\footnote{\url{https://www.ap.org/en-gb/}.} news articles from 2009
to 2016.  \imdb is a set of movie reviews collected by
\citet{Maas+:2011}. \bnc is the written portion of the British National
Corpus \citep{bnc07}, which contains excerpts from journals, books,
letters, essays, memoranda, news and other types of text. For \apnews
and \bnc, we randomly sub-sample a set of documents for our experiments.

For preprocessing, we tokenise words and sentences using Stanford
CoreNLP \citep{Klein:Manning:2003}. We lowercase all word tokens, filter
word types that occur less than 10 times, and exclude the top 0.1\% most
frequent word types.\footnote{For the topic model, we remove word tokens
  that correspond to these filtered word types; for the language model
  we represent them as \texttt{\unk} tokens (as for unseen words in test).}
We additionally remove stopwords for the topic model document
context.\footnote{We use Mallet's stopword list:
  \url{https://github.com/mimno/Mallet/tree/master/stoplists}.}  All
datasets are partitioned into training, development and test sets;
preprocessed dataset statistics are presented in \tabref{dataset}.

\begin{table*}
\begin{center}
\begin{adjustbox}{max width=0.6\textwidth}
\begin{tabular}{cc@{\;}ccc@{\;}ccc@{\;}c}
\toprule
\multirow{2}{*}{\textbf{Collection}} & \multicolumn{2}{c}{\textbf{Training}}  
&& \multicolumn{2}{c}{\textbf{Development}} &&
\multicolumn{2}{c}{\textbf{Test}} \\
  \cmidrule{2-3}
  \cmidrule{5-6}
  \cmidrule{8-9}
& \#Docs & \#Tokens && \#Docs & \#Tokens && \#Docs & \#Tokens \\
\midrule
\apnews & 50K & 15M && 2K & 0.6M && 2K & 0.6M \\
\imdb & 75K & 20M && 12.5K & 0.3M && 12.5K & 0.3M \\
\bnc & 15K & 18M && 1K & 1M && 1K & 1M \\
\bottomrule
\end{tabular}
\end{adjustbox}
\end{center}
\caption{Preprocessed dataset statistics.}
\label{tab:dataset}
\end{table*}

\begin{table*}[t]
\begin{center}
\begin{tabular}{ccccc@{\;\;}c@{\;\;}cc@{\;\;}c@{\;\;}c@{\;\;}c}
\toprule
\multirow{2}{*}{\textbf{Domain}} & \multirow{2}{*}{\textbf{LSTM Size}} &  
{\textbf{\method{vanilla-}}} & \multirow{2}{*}{\textbf{\lclm}} & 
\multicolumn{3}{c}{\textbf{\lstmlda}} &&
\multicolumn{3}{c}{\textbf{\tdlm}} \\
\cmidrule{5-7}
\cmidrule{9-11}
&&\textbf{\method{lstm}}&& 50 & 100 & 150 && 50 & 100 & 150 \\
\midrule
\multirow{2}{*}{\apnews} & small &  64.13 & 54.18 & 57.05 & 55.52 & 
54.83 && 53.00 & 52.75 & \textbf{52.65} \\
& large & 58.89 & 50.63 & 52.72 & 50.75 & 50.17 && 48.96 & 48.97 & 
\textbf{48.21} \\
\multirow{2}{*}{\imdb} & small & 72.14 & 67.78 &  69.58 & 69.64 & 69.62 
&& 63.67 & \textbf{63.45} & 63.82 \\
& large & 66.47 & 67.86 & 63.48 & 63.04 & 62.78 && 58.99 & 59.04 & 
\textbf{58.59} \\
\multirow{2}{*}{\bnc} & small & 102.89\z & 87.47 & 96.42 & 96.50 & 96.38 &&
87.42 & \textbf{85.99} & 86.43 \\
& large & 94.23 & 80.68 & 88.42 & 87.77 & 87.28 && 82.62 & 81.83 & 
\textbf{80.58} \\
\bottomrule
\end{tabular}
\end{center}
\caption{Language model perplexity performance of all models over 
\apnews, \imdb and \bnc. Boldface indicates best performance in each 
row.}
\label{tab:lm}
\end{table*}

We tune hyper-parameters of \tdlm based on development set language 
model perplexity. In general, we find that optimal settings are fairly 
robust across collections, with the exception of $m_3$, as document length is 
collection dependent; optimal hyper-parameter values are given in 
\tabref{hyperparameters}.  In terms of LSTM size, we explore 
2 settings: a small model with 1 LSTM layer and 600 hidden 
units, and a large model with 2 layers and 900 hidden 
units.\footnote{Multi-layer LSTMs are vanilla stacked LSTMs without skip 
connections \citep{Gers+:2000} or depth-gating \citep{Yao+:2015}.} For 
the topic number, we experiment with 50, 100 and 150 topics. Word 
embeddings are pre-trained $300$-dimension \wordtovec Google News 
vectors.\footnote{\url{https://code.google.com/archive/p/word2vec/}.}

For comparison, we compare \tdlm with:\footnote{Note that all models use the same pre-trained
  \wordtovec vectors.}
\paragraph{\textbf{\lstm}:}
A standard LSTM language model, using the same \tdlm hyper-parameters 
where applicable. This is the baseline model.

\paragraph{\textbf{\lclm}:}
A larger context language model that incorporates context from preceding
sentences \citep{Wang+:2016}, by treating the preceding sentence as a
bag of words, and using an attentional mechanism when predicting the
next word. An additional hyper-parameter in \lclm is the number of
preceeding sentences to incorporate, which we tune based on a
development set (to 4 sentences in each case). All other
hyper-parameters (such as $n_\mathit{batch}$, $e$, $n_\mathit{epoch}$,
$k_2$) are the same as \tdlm.

\paragraph{\textbf{\lstmlda}:}
A standard LSTM language model that incorporates LDA topic information.
We first train an LDA model \citep{Blei+:2003,Griffiths:2004} to learn
50/100/150 topics for \apnews, \imdb and \bnc.\footnote{Based on Gibbs
  sampling; $\alpha = 0.1$, $\beta = 0.01$.} For a document, the LSTM
incorporates the LDA topic distribution ($\mathbf{q}$) by concatenating
it with the output hidden state ($\mathbf{h}_t$) to predict the next
word (i.e.\ $\mathbf{h}'_t = \mathbf{h}_t \oplus \mathbf{q}$).  That is,
it incorporates topical information into the language model, but unlike
\tdlm the language model and topic model are trained separately.

We present language model perplexity performance in \tabref{lm}. All
models outperform the baseline \lstm, with \tdlm performing the best
across all collections. \lclm is competitive over the \bnc, although the
superiority of \tdlm for the other collections is substantial.
\lstmlda performs relatively well over \apnews and \imdb, but very
poorly over \bnc.

The strong performance of \tdlm over \lclm suggests that compressing
document context into topics benefits language modelling more than using
extra context words directly.\footnote{The context size of \lclm
  (4 sentences) is technically smaller than \tdlm (full document),
  however, note that increasing the context size does not benefit \lclm,
  as the context size of 4 gives the best performance.} Overall, our
results show that topical information can help language modelling and
that joint inference of topic and language model produces the best
results.

\begin{figure*}[t]
        \centering
        \begin{subfigure}[b]{0.3\textwidth}
                \includegraphics[width=\textwidth]{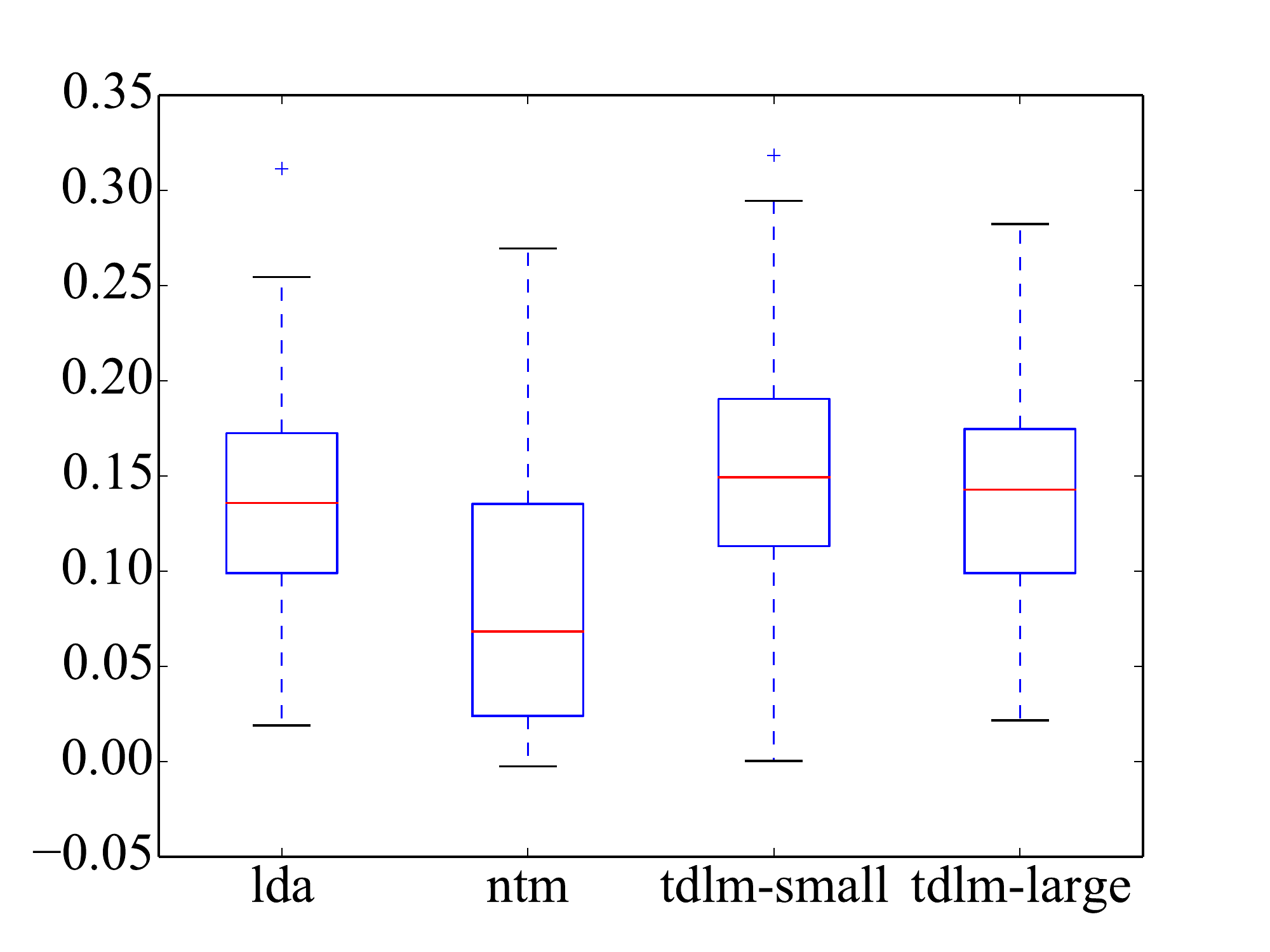}
                \caption{\apnews}
        \end{subfigure}
        ~
        \begin{subfigure}[b]{0.3\textwidth}
                \includegraphics[width=\textwidth]{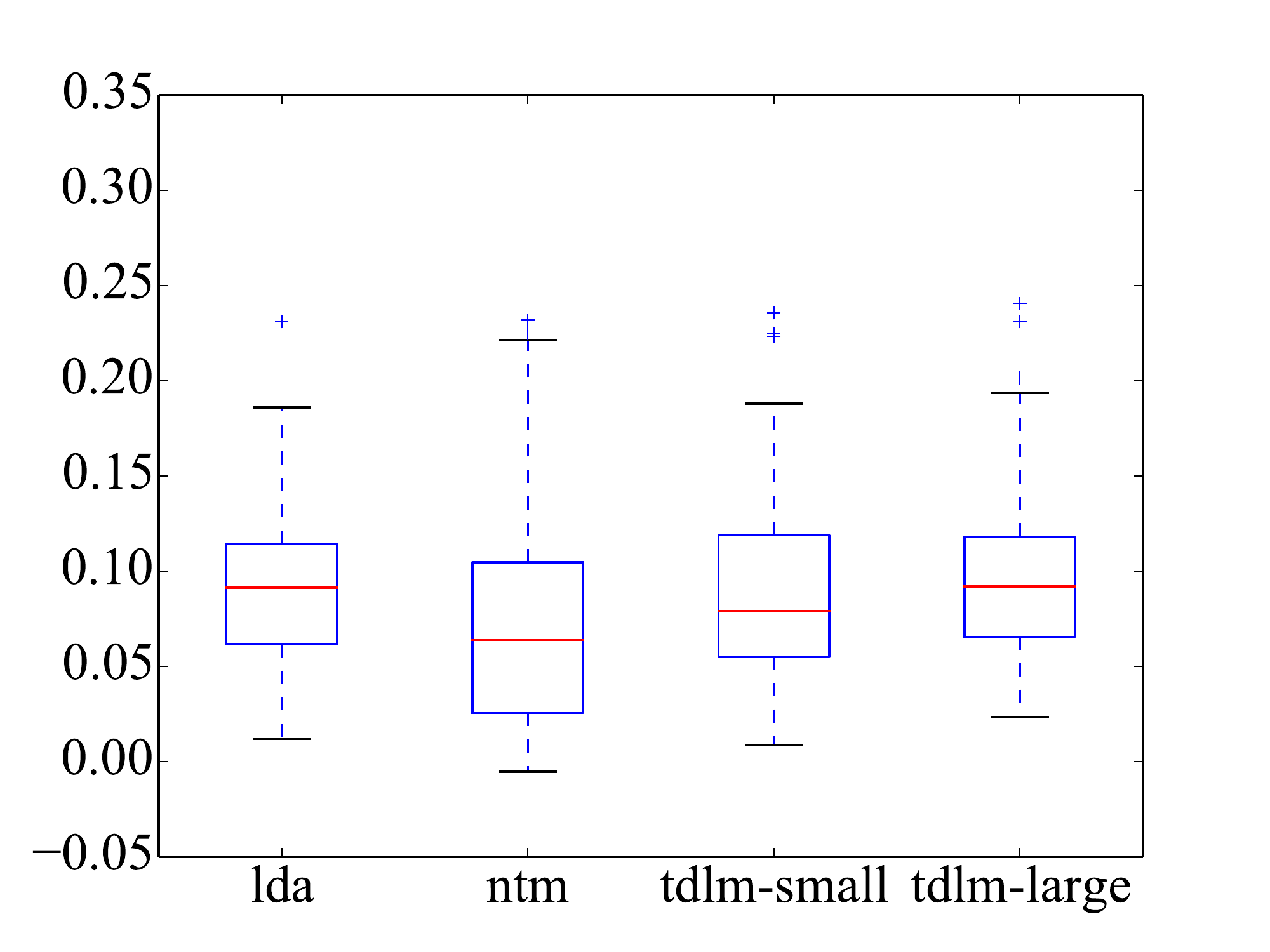}
                \caption{\imdb}
        \end{subfigure}
        ~
        \begin{subfigure}[b]{0.3\textwidth}
                \includegraphics[width=\textwidth]{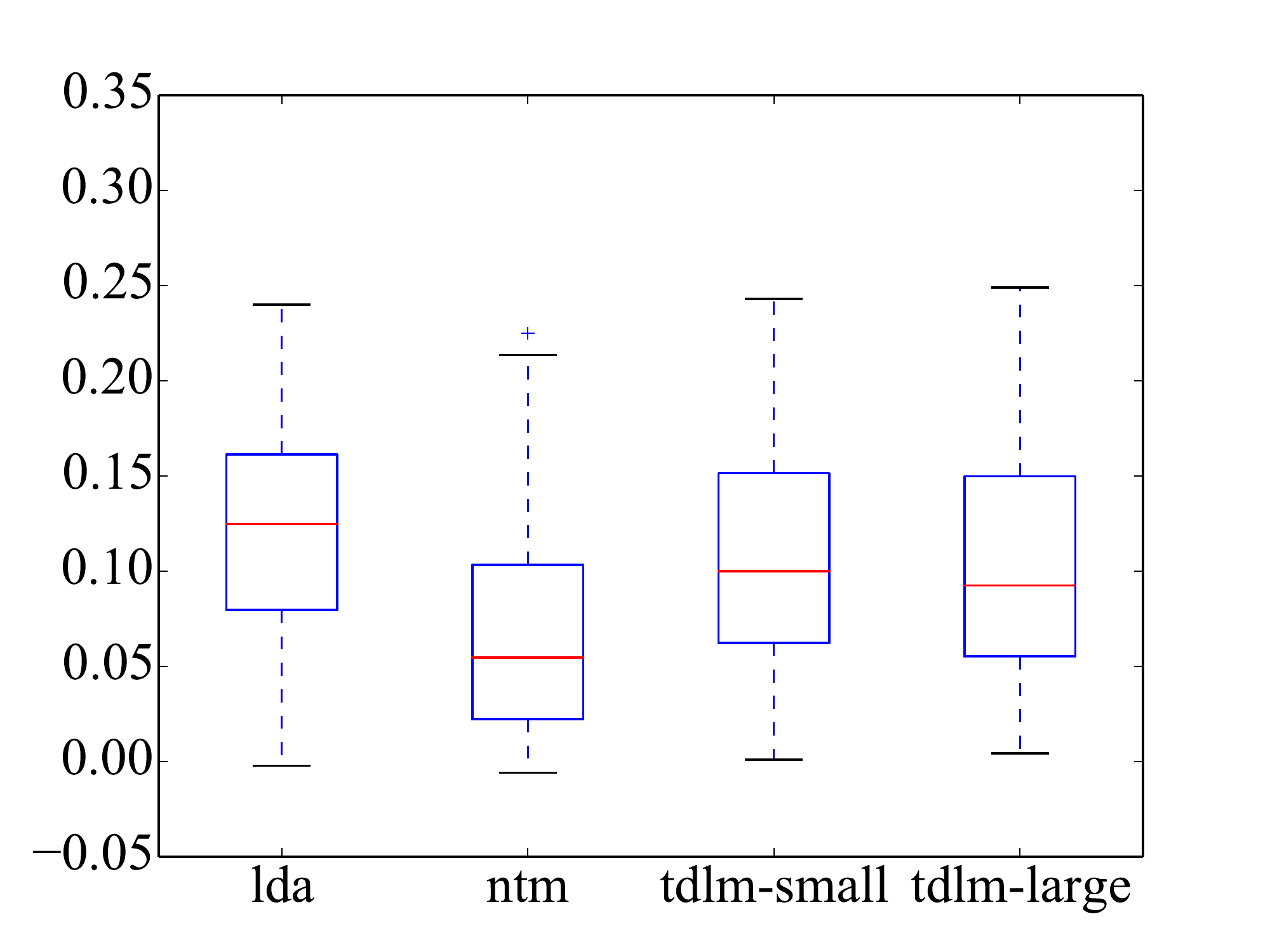}
                \caption{\bnc}
        \end{subfigure}
\caption{Boxplots of topic coherence of all models; number of topics $= 
100$.}
\label{fig:boxplot}
\end{figure*}

\section{Topic Model Evaluation}
\label{sec:tm}

We saw that \tdlm performs well as a language model, but it is also a
topic model, and like LDA it produces: (1) a probability distribution
over topics for each document (\eqnref{attn}); and (2) a probability
distribution over word types for each topic.

Recall that $\mathbf{s}$ is a weighted mean of topic vectors for a 
document (\eqnref{hiddenreps}).  Generating the vocabulary distribution 
for a particular topic is therefore trivial: we can do so by treating 
$\mathbf{s}$ as having maximum weight (1.0) for the topic of interest, 
and no weight (0.0) for all other topics. Let $\mathbf{B}_t$ denote the 
topic output vector for the \mbox{$t$-th} topic. To generate the 
multinomial distribution over word types for the $t$-th topic, we 
replace $\mathbf{s}$ with $\mathbf{B}_t$ before computing the softmax 
over the vocabulary.

Topic models are traditionally evaluated using model perplexity.  There
are various ways to estimate test perplexity \citep{Wallach:2009}, but
\citet{Chang+:09} show that perplexity does not correlate with the
coherence of the generated topics.
\citet{Newman+:2010a,Mimno+:2011,Aletras+:2013} propose automatic
approaches to computing topic coherence, and \citet{Lau+:2014}
summarises these methods to understand their differences. We propose
using automatic topic coherence as a means to evaluate the topic model
aspect of \tdlm.

Following \citet{Lau+:2014}, we compute topic coherence using normalised
PMI (``NPMI'') scores.  Given the top-$n$ words of a topic, coherence is 
computed based on the sum of pairwise NPMI scores between topic words, 
where the word probabilities used in the NPMI calculation are based on 
co-occurrence statistics mined from English Wikipedia with a sliding 
window \citep{Newman+:2010a,Lau+:2014}.\footnote{We use this toolkit to 
compute topic coherence: 
\url{https://github.com/jhlau/topic_interpretability}.}

Based on the findings of \citet{Lau:Baldwin:2016a}, we average topic
coherence over the top-5/10/15/20 topic words.  To aggregate topic
coherence scores for a model, we calculate the mean coherence over
topics.

In terms of datasets, we use the same document collections (\apnews, 
\imdb and \bnc) as the language model experiments (\secref{lm}).  We use 
the same hyper-parameter settings for \tdlm and do not tune them.

For comparison, we use the following topic models:

\paragraph{\textbf{\lda}:}
We use a LDA model as a baseline topic model. We use the same LDA models
as were used to learn topic distributions for \lstmlda (\secref{lm}).

\paragraph{\textbf{\ntm}:}
\ntm is a neural topic model proposed by \citet{Cao+:2015}. The 
document-topic and topic-word multinomials are expressed from a neural 
network perspective using differentiable functions. Model 
hyper-parameters are tuned using development loss.

\begin{table}[t]
\begin{center}
\begin{adjustbox}{max width=0.45\textwidth}
\begin{tabular}{ccccc}
\toprule
\multirow{2}{*}{\textbf{Topic No.}} & \multirow{2}{*}{\textbf{System}} & 
\multicolumn{3}{c}{\textbf{Coherence}} \\
&& \apnews & \imdb & \bnc \\
\midrule
\multirow{4}{*}{50} & \lda & .125 & .084 & \textbf{.106} \\
& \ntm & .075 & .064  & .081 \\
& \tdlms & \textbf{.149} & \textbf{.104} & .102\\
& \tdlml & .130 & .088 & .095 \\
\hdashline
\multirow{4}{*}{100} & \lda & .136 & .092 & \textbf{.119} \\
& \ntm & .085 & .071 & .070 \\
& \tdlms & \textbf{.152} & .087 & .106 \\
& \tdlml & .142 & \textbf{.097} &  .101 \\
\hdashline
\multirow{4}{*}{150} & \lda & .134 & \textbf{.094} & \textbf{.119} \\
& \ntm & .078 & .075 &  .072 \\
& \tdlms & \textbf{.147} & .085 & .100 \\
& \tdlml & .145 & .091 & .104 \\
\bottomrule
\end{tabular}
\end{adjustbox}
\end{center}
\caption{Mean topic coherence of all models over \apnews, \imdb and 
  \bnc. Boldface indicates the best performance for each dataset and topic 
  setting.}
\label{tab:tm}
\end{table}

Topic model performance is presented in \tabref{tm}. There are two
models of \tdlm (\tdlms and \tdlml), which specify the size of its LSTM
model (1 layer$+$600 hidden vs.\ 2 layers$+$900 hidden; see
\secref{lm}).  \tdlm achieves encouraging results: it has the best
performance over \apnews, and is competitive over \imdb.  \lda, however,
produces more coherent topics over \bnc.  Interestingly, coherence
appears to increase as the topic number increases for \lda, but the
trend is less pronounced for \tdlm.  \ntm performs the worst of the 3
topic models, and manual inspection reveals that topics are in general 
not very interpretable.  Overall, the results suggest that \tdlm topics 
are competitive: at best they are more coherent than \lda topics, and at 
worst they are as good as \lda topics.

To better understand the spread of coherence scores and impact of 
outliers, we present box plots for all models (number of topics $=100$) 
over the 3 domains in \figref{boxplot}. Across all domains, \ntm has 
poor performance and larger spread of scores. The difference between 
\lda and \tdlm is small (\tdlm $>$ \lda in \apnews, but \lda $<$ \tdlm 
in \bnc), which is consistent with our previous observation that \tdlm topics are 
competitive with \lda topics.

\section{Extensions}

One strength of \tdlm is its flexibility, owing to it taking the form of
a neural network. To showcase this flexibility, we explore two simple
extensions of \tdlm, where we: (1) build a supervised model using
document labels (\secref{supervised}); and (2) incorporate additional
document metadata (\secref{metadata}).

\subsection{Supervised Model}
\label{sec:supervised}

\begin{table}[t]
\footnotesize
\begin{center}
\begin{tabular}{ccc}
\toprule
\textbf{Partition} & \textbf{\#Docs} & \textbf{\#Tokens} \\
\midrule
Training & 9314 & 2.6M \\
Development & 2000 & 0.5M \\
Test & 7532 & 1.7M \\
\bottomrule
\end{tabular}
\end{center}
\caption{\tnews preprocessed statistics.}
\label{tab:20news}
\end{table}

In datasets where document labels are known, supervised topic model 
extensions are designed to leverage the additional information to 
improve modelling quality. The supervised setting also has an additional 
advantage in that model evaluation is simpler, since models can be 
quantitatively assessed via classification accuracy.

To incorporate supervised document labels, we treat document 
classification as another sub-task in \tdlm.  Given a document and its 
label, we feed the document through the topic model network to generate 
the document vector $\mathbf{d}$ and document-topic representation 
$\mathbf{s}$, and then we concatenate both and connect it to another 
dense layer with softmax output to generate the probability distribution 
over classes.

During training, we have additional minibatches for the documents. We 
start the document classification training after the topic and language 
models have completed training in each epoch.

We use \tnews in this experiment, which is a popular dataset for text
classification.  \tnews is a collection of forum-like messages from 20
newsgroups categories. We use the ``bydate'' version of the dataset,
where the train and test partition is separated by a specific date.  We
sample 2K documents from the training set to create the development
set. For preprocessing we tokenise words and sentence using Stanford
CoreNLP \citep{Klein:Manning:2003}, and lowercase all words. As with
previous experiments (\secref{lm}) we additionally filter low/high
frequency word types and stopwords.  Preprocessed dataset statistics are
presented in \tabref{20news}.

For comparison, we use the same two topic models as in \secref{tm}: \ntm
and \lda.  Both \ntm and \lda have natural supervised extensions
\citep{Cao+:2015,Blei+:2008} for incorporating document labels.  For
this task, we tune the model hyper-parameters based on development
accuracy.\footnote{Most hyper-parameter values for \tdlm are similar to
  those used in the language and topic model experiments; the only
  exceptions are: $a = 80$, $b= 100$, $n_\mathit{epoch}=20$,
  $m_3=150$. The increase in parameters is unsurprising, as the
  additional supervision provides more constraint to the model.}
Classification accuracy for all models is presented in
\tabref{classification}. We present \tdlm results using only the small
setting of LSTM (1 layer $+$ 600 hidden), as we found there is little
gain when using a larger LSTM.

\ntm performs very strongly, outperforming both \lda and \tdlm by a
substantial margin.  Comparing \lda and \tdlm, \tdlm achieves better
performance, especially when there is a smaller number of topics.  Upon
inspection of the topics we found that \ntm topics are much less
coherent than those of \lda and \tdlm, consistent with our observations from
\secref{tm}. 

\begin{table}[t]
\footnotesize
\begin{center}
\begin{tabular}{ccc}
\toprule
{\textbf{Topic No.}} & {\textbf{System}} & \textbf{Accuracy} \\
\midrule
\multirow{3}{*}{50} & \lda & .567  \\
& \ntm &  \textbf{.649} \\
& \tdlm & .606 \\
\hdashline
\multirow{3}{*}{100} & \lda & .581  \\
& \ntm &  \textbf{.639} \\
& \tdlm & .602 \\
\hdashline
\multirow{3}{*}{150} & \lda & .597 \\
& \ntm &  \textbf{.628} \\
& \tdlm & .601 \\
\bottomrule
\end{tabular}
\end{center}
\caption{\tnews classification accuracy. All models are supervised 
extensions of the original models. Boldface indicates the best 
performance for each topic setting.}
\label{tab:classification}
\end{table}
  
\begin{table}[t]
\begin{center}
\begin{adjustbox}{max width=0.45\textwidth}
\begin{tabular}{cccc}
\toprule
\textbf{Topic No.} & \textbf{Metadata} & \textbf{Coherence} & 
\textbf{Perplexity} \\
\midrule
\multirow{2}{*}{50} & No & .128 & 52.45 \\
& Yes & \textbf{.131} & \textbf{51.80} \\
\hdashline
\multirow{2}{*}{100} & No & \textbf{.142} & 52.14 \\
& Yes & .139 & \textbf{51.76} \\
\hdashline
\multirow{2}{*}{150} & No & .135 & 52.25 \\
& Yes & \textbf{.143} & \textbf{51.58} \\
\bottomrule
\end{tabular}
\end{adjustbox}
\end{center}
\caption{Topic coherence and language model perplexity by incorporating 
classification tags on \apnews. Boldface indicates optimal coherence and 
perplexity performance for each topic setting.}
\label{tab:tag}
\end{table}

\subsection{Incorporating Document Metadata}
\label{sec:metadata}

In \apnews, each news article contains additional document metadata,
including subject classification tags, such as ``General News'',
``Accidents and Disasters'', and ``Military and Defense''.  We present
an extension to incorporate document metadata in \tdlm to demonstrate
its flexibility in integrating this additional information.

As some of the documents in our original \apnews sample were missing
tags, we re-sampled a set of \apnews articles of the same size as our
original, all of which have tags. In total, approximately 1500 unique
tags can be found among the training articles.

To incorporate these tags, we represent each of them as a learnable 
vector and concatenate it with the document vector before computing the 
attention distribution. Let $\mathbf{z}_i \in \mathbb{R}^f$ denote the 
$f$-dimension vector for the $i$-th tag. For the $j$-th document, we sum 
up all tags associated with it:
\begin{equation*}
 {\mathbf{e}} = \sum^{n_{tags}}_{i=1} \mathbb{I}(i,j) \mathbf{z}_i
\end{equation*}
where $n_{tags}$ is the total number of unique tags, and function 
$\mathbb{I}(i,j)$ returns 1 is the $i$-th tag is in the $j$-th document 
or 0 otherwise. We compute $\mathbf{d}$ as before (\secref{tmc}), and 
concatenate it with the summed tag vector: $\mathbf{d}' = \mathbf{d} 
\oplus {\mathbf{e}}$.

\begin{figure}[t]
\begin{center}
\includegraphics[width=0.48\textwidth]{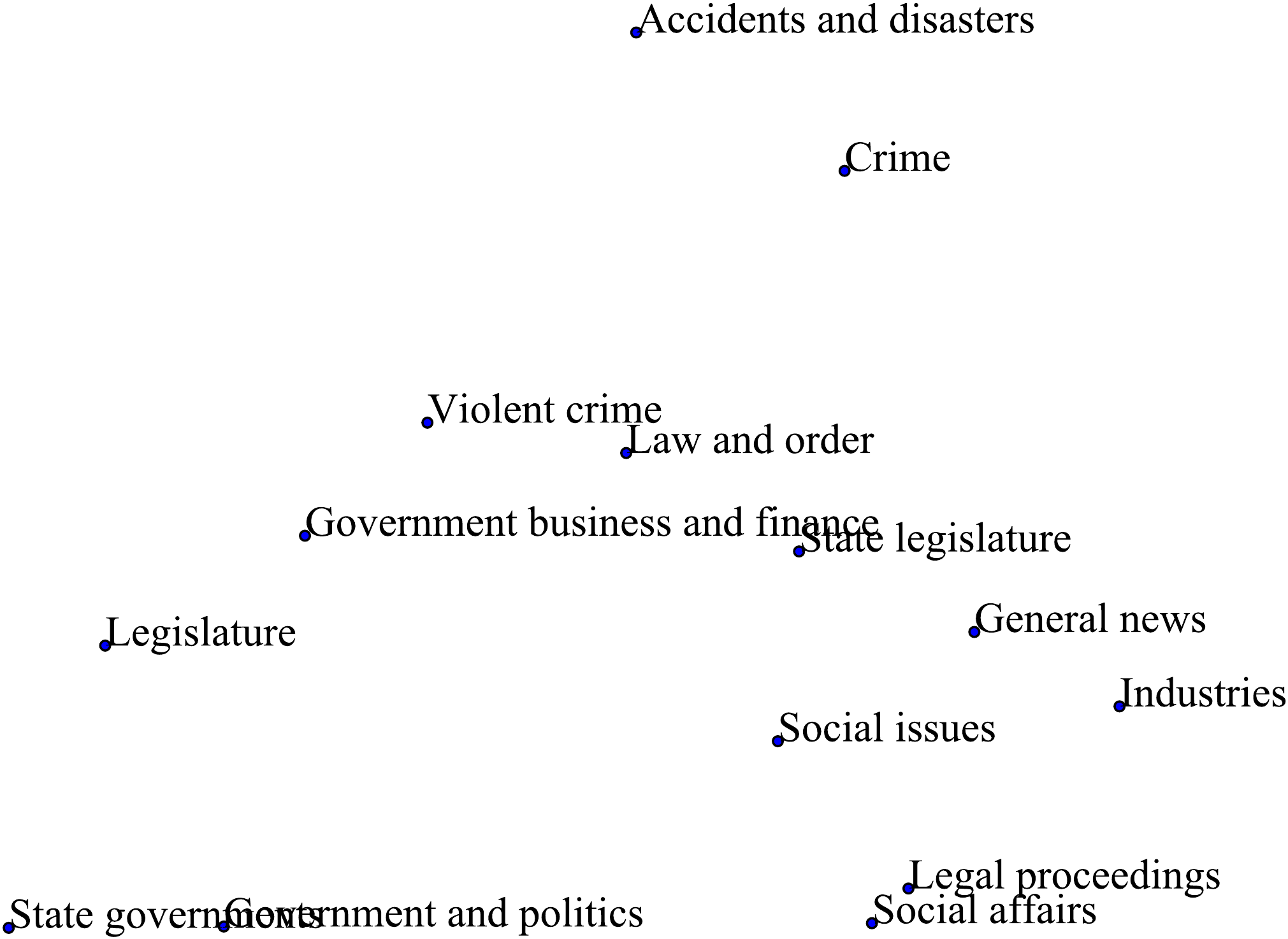}
\end{center}
\caption{Scatter plots of tag embeddings (model=150 topics)}
\label{fig:tagemb}
\end{figure}

\begin{table*}
\begin{center}
\begin{adjustbox}{max width=1.0\textwidth}
\begin{tabular}{lp{17.05cm}}
\toprule
\textbf{Topic} & \textbf{Generated Sentences} \\
\midrule
\multirow{3}{*}{\begin{tabular}[c]{@{}l@{}l@{}}protesters suspect gunman 
\\officers occupy gun arrests \\ suspects shooting officer\end{tabular}} 
& $\bullet$ police say a suspect in the shooting was shot in the chest 
and later shot and killed by a police officer .  \\
& $\bullet$ a police officer shot her in the chest and the man was 
killed . \\
& $\bullet$ police have said four men have been killed in a shooting in 
suburban london . \\
\hdashline
\multirow{4}{*}{\begin{tabular}[c]{@{}l@{}l@{}}film awards actress  
comedy \\ music actor album  show \\ nominations movie\end{tabular}} & 
$\bullet$ it 's like it 's not fair to keep a star in a light , " he 
says . \\
& $\bullet$ but james , a four-time star , is just a \unk . \\
& $\bullet$ a \unk adaptation of the movie " the dark knight rises 
" won best picture and he was nominated for best drama for best director 
of " \unk , " which will be presented sunday night . \\
\hdashline
\multirow{4}{*}{\begin{tabular}[c]{@{}l@{}l@{}}storm snow weather  
inches \\ flooding rain  service \\ winds tornado  
forecasters\end{tabular}} &
$\bullet$ temperatures are forecast to remain above freezing enough to 
reach a tropical storm or heaviest temperatures .\\
& $\bullet$ snowfall totals were one of the busiest in the country . \\
& $\bullet$ forecasters say tornado irene 's strong winds could ease 
visibility and funnel clouds of snow from snow monday to the mountains 
.\\
\hdashline
\multirow{3}{*}{\begin{tabular}[c]{@{}l@{}l@{}} virus nile flu vaccine 
\\ disease outbreak infected \\ symptoms cough tested \end{tabular}} &
$\bullet$ he says the disease was transmitted by an infected person . \\
& $\bullet$ \unk says the man 's symptoms are spread away from the 
heat .  \\
& $\bullet$ meanwhile in the \unk , the virus has been common in 
the mojave desert . \\
\bottomrule
\end{tabular}
\end{adjustbox}
\end{center}
\caption{Generated sentences for \apnews topics.}
\label{tab:sample}
\end{table*}

We train two versions of \tdlm on the new \apnews dataset: (1) the
vanilla version that ignores the tag information; and (2) the extended
version which incorporates tag information.\footnote{Model
  hyper-parameters are the same as the ones used in the language
  (\secref{lm}) and topic model (\secref{tm}) experiments.} We
experimented with a few values for the tag vector size ($f$) and find
that a small value works well; in the following experiments we use
$f=5$. We evaluate the models based on language model perplexity and
topic model coherence, and present the results in
\tabref{tag}.\footnote{As the vanilla \tdlm is trained on the new
  \apnews dataset, the numbers are slightly different to those in
  \tabref[s]{lm} and \ref{tab:tm}.}

In terms of language model perplexity, we see a consistent improvement 
over different topic settings, suggesting that the incorporation of tags 
improves modelling.  In terms of topic coherence, there is a small but 
encouraging improvement (with one exception).

To investigate whether the vectors learnt for these tags are meaningful,
we plot the top-$14$ most frequent tags in \figref{tagemb}.\footnote{The
  5-dimensional vectors are compressed using PCA.} The plot seems
reasonable: there are a few related tags that are close to each other,
e.g.\ ``State government'' and ``Government and politics''; ``Crime''
and ``Violent Crime''; and ``Social issues'' and ``Social affairs''.

\section{Discussion}

Topics generated by topic models are typically interpreted by way of
their top-$N$ highest probability words. In \tdlm, we can additionally
generate sentences related to the topic, providing another way to
understand the topics. To do this, we can constrain the topic vector for
the language model to be the topic output vector of a particular topic
(\eqnref{incorporatetm}).

We present 4 topics from a \apnews model ($k=100$; LSTM size $=$
``large'') and 3 randomly generated sentences conditioned on each topic
in \tabref{sample}.\footnote{Words are sampled with temperature
  $= 0.75$. Generation is terminated when a special end symbol is
  generated or when sentence length is greater than 40 words.} The
generated sentences highlight the content of the topics, providing
another interpretable aspect for the topics. These results also 
reinforce that the language model is driven by topics.



\section{Conclusion}

We propose \tdlm, a topically driven neural language model. \tdlm has 
two components: a language model and a topic model, which are jointly
trained using a neural network. We demonstrate that \tdlm outperforms a
state-of-the-art language model that incorporates larger context, and
that its topics are potentially more coherent than LDA topics. We 
additionally propose simple extensions of \tdlm to incorporate 
information such as document labels and metadata, and achieved 
encouraging results.

\section*{Acknowledgments}

We thank Shraey Bhatia for providing an open source implementation of 
\ntm, and the anonymous reviewers for their insightful comments and 
valuable suggestions. This work was funded in part by the Australian
Research Council.

\bibliographystyle{acl_natbib}
\bibliography{strings,citations,papers}

\end{document}